\documentclass[runningheads]{llncs}
\usepackage{graphicx}
\usepackage{comment}
\usepackage{amsmath,amssymb} 
\usepackage{color}
\usepackage{enumitem}
\usepackage{multirow}
\usepackage{booktabs}
\usepackage{enumerate,enumitem}

\newcommand{\R}{\mathbb{R}}
\setlist[itemize]{noitemsep, nolistsep}

\begin{document}
\pagestyle{headings}
\mainmatter
\def\ECCVSubNumber{3892}  

\title{Pillar-based Object Detection \\ for Autonomous Driving} 

%
\author{Yue Wang\inst{1,2} \and
Alireza Fathi \inst{2} \and
Abhijit Kundu \inst{2} \and
David Ross \inst{2} \and \\
Caroline Pantofaru \inst{2} \and
Thomas Funkhouser \inst{2} \and
Justin Solomon \inst{1}
}
\authorrunning{Wang et al.}
%
\institute{MIT \\
\email{\{yuewangx,jsolomon\}@mit.edu} \and
Google \\
\email{\{alirezafathi,abhijitkundu,dross,cpantofaru,tfunkhouser\}@google.com}}
\maketitle

\begin{abstract}
    We present a simple and flexible object detection framework optimized for autonomous driving. Building on the observation that point clouds in this application are extremely sparse, we propose a practical \emph{pillar-based} approach to fix the imbalance issue caused by anchors. In particular, our algorithm incorporates a cylindrical projection into multi-view feature learning, predicts bounding box parameters per pillar rather than per point or per anchor, and includes an aligned pillar-to-point projection module to improve the final prediction. Our anchor-free approach avoids hyperparameter search associated with past methods, simplifying 3D object detection while significantly improving upon state-of-the-art. 
    %
\end{abstract}
\section{Introduction}

3D object detection is a central component of perception systems for autonomous driving, used to identify pedestrians, vehicles, obstacles, and other key features of the environment around a car. Ongoing development of object detection methods for vision, graphics, and other domain areas has led to steady improvement in the performance and reliability of these systems as they transition from research to production.

Most 3D object detection algorithms project points to a single prescribed view for feature learning.  These views---e.g., the ``bird's eye'' view of the environment around the car---are not necessarily optimal for distinguishing objects of interest. 
After computing features, these methods typically make \emph{anchor-based} predictions of the locations and poses of objects in the scene. Anchors provide useful position and pose priors, but they lead to learning algorithms with many hyperparameters and potentially unstable training.  


Two popular architectures typifying this approach 
are PointPillars~\cite{Lang_2019_CVPR} and multi-view fusion (MVF)~\cite{Zhou2019EndtoEndMF}, which achieve top efficiency and performance on recent 3D object detection benchmarks.  These methods use learning representations built from birds-eye view pillars above the ground plane. 
MVF also benefits from complementary information provided by a spherical view.
These methods, however, predict parameters of a bounding box per anchor. Hyperparameters of anchors need to be tuned case-by-case for different tasks/datasets, reducing practicality. Moreover, anchors are sparsely distributed in the scene, leading to a significant class imbalance. An anchor is assigned as positive when its intersection-over-union (IoU) with a ground-truth box reaches above prescribed threshold;  the number of positive anchors is less than 0.1\% of all anchors in a typical point cloud. 

As an alternative, we introduce a fully pillar-based (anchor-free) object detection model for autonomous driving. In principle, our method is an intuitive extension of PointPillars~\cite{Lang_2019_CVPR} and MVF~\cite{Zhou2019EndtoEndMF} that uses pillar representations in multi-view feature learning and in pose estimation. In contrast to past works, we find that predicting box parameters per anchor is neither necessary nor effective for 3D object detection in autonomous driving. A critical new component of our model is a per-pillar prediction network, removing the necessity of anchor assignment. For each birds-eye view pillar, the model directly predicts the position and pose of the best possible box.  
This component improves performance and is significantly simpler than current state-of-the-art 3D object detection pipelines. 

In addition to introducing this pillar-based object detection approach, we also propose ways to address other problems with previous methods.   For example, we find the spherical projection in MVF~\cite{Zhou2019EndtoEndMF} causes unnecessary distortion of scenes and can actually degrade detection performance. So, we complement the conventional birds-eye view with a new cylindrical view, which does not suffer from perspective distortions.
We also observe that current methods for pillar-to-point projection suffer from spatial aliasing, which we improve with bilinear interpolation.

To investigate the performance of our method, we train and test the model on the Waymo Open Dataset~\cite{Sun2019ScalabilityIP}. Compared to the top performers on this dataset~\cite{Ngiam2019StarNetTC,Zhou2019EndtoEndMF,Lang_2019_CVPR}, we show significant improvements by 6.87 3D mAP and 6.71 2D mAP for vehicle detection.
We provide ablation studies to analyze the contribution of each proposed module in \S\ref{sec:exp} and show that each outperforms its counterpart by a large margin.

\paragraph*{Contributions.} We summarize our key contributions as follows:
\begin{itemize}

  \item  We present a fully pillar-based model for high-quality 3D object detection. The model achieves state-of-the-art results on the most challenging autonomous driving dataset. 

  
  \item We design an pillar-based box prediction paradigm for object detection, which is much simpler and stronger than its anchor-based and/or point-based counterpart. 
  
  \item We analyze the multi-view feature learning module and find a cylindrical view is the best complementary view to a birds-eye view. 
    
  \item We use bilinear interpolation in pillar-to-point projection to avoid quantization errors. 

  
  \item We release our code to facilitate reproducibility and future research:\\ \small{\url{https://github.com/WangYueFt/pillar-od}}. 
\end{itemize}
\section{Related Work}
\label{sec:related_work}

Methods for object detection are highly successful in 2D visual recognition~\cite{girshick2014rcnn,girshick15fastrcnn,ren2015faster,he2017maskrcnn,liu2016ssd,redmonF17}. They generally involve two aspects: backbone networks and detection heads. The input image is passed through a backbone network to learn latent features, while the detection heads make predictions of bounding boxes based on the features. 
In 3D, due to the sparsity of the data, many special considerations are taken to improve both efficiency and performance. Below, we discuss related works on 
general object detection, as well as more general methods relevant to learning on point clouds.

\textit{2D object detection.} RCNN~\cite{girshick2014rcnn} pioneers the modern two-stage approach to object detection; more recent models often follow a similar template. RCNN uses a simple selective search to find regions of interest (region proposals) and subsequently applies a convolutional neural network (CNN) to bottom-up region proposals to regress bounding box parameters. 

RCNN can be inefficient because it applies a CNN to each region proposal, or image patch. Fast RCNN~\cite{girshick15fastrcnn} addresses this problem by sharing features for region proposals from the same image: 
it passes the image in a one-shot fashion through the CNN, and then region features are cropped and resized from the shared feature map. Faster RCNN~\cite{ren2015faster} further improves speed and performance by replacing the selective search with region proposal networks (RPN), whose features 
can be shared. 

Mask RCNN~\cite{he2017maskrcnn} is built on top of Faster RCNN. In addition to box prediction, it adds another pathway for mask prediction, enabling enables object detection, semantic segmentation, and instance segmentation using a single pipeline. Rather than using ROIPool~\cite{girshick15fastrcnn} to resize feature patch to a fixed size grid, Mask RCNN proposes using bilinear interpolation (ROIAlign) to avoid quantization error. 
Beyond significant structural changes in the general two-stage object detection models, extensions using machinery from image processing and shape registration include: exploiting multi-scale information using feature pyramids~\cite{Lin2017}, iterative refinement of box prediction~\cite{cai18cascadercnn}, and using deformable convolutions~\cite{dai17dcn} to get an adaptive receptive field. Recent works~\cite{Zhu2019FeatureSA,Tian2019FCOSFC,Zhou2019ObjectsAP} also show anchor-free  methods achieve comparable results to existing two-stage object detection models in 2D. 

In addition to two-stage object detection, many works aim to design real-time object detection models via one-stage algorithms. These methods densely place anchors that define position priors and size priors in the image and then associate each anchor with the ground-truth using an intersection-over-union (IoU) threshold. The networks classify each anchor and regress 
parameters of anchors; non-maximum suppression (NMS) removes redundant predictions.  SSD~\cite{liu2016ssd} and YOLO~\cite{Redmon2015YouOL,redmonF17} are representative examples of this approach. RetinaNet~\cite{lin2017focal} is built on the observation that the extreme foreground-background class imbalance encountered during training causes one-stage detectors trailed the accuracy of two-stage detectors. 
It proposes a focal loss to amplify a sparse set of hard examples and to prevent easy negatives from overwhelming the detector during training. 
Similar to image object detection, we also find the imbalance issue causes instability in 3D object detection. In contrast to RetinaNet, however, we replace anchors with pillar-centric predictions to alleviate imbalance. 

\textit{Learning on point clouds.} Point clouds provide a natural representation of 3D shapes~\cite{Chang2015ShapeNetAI} and scenes. Due to  irregularity and symmetry under reordering, however, defining convolution-like operations on point clouds is difficult. 

PointNet~\cite{qi2017pointnet} exemplifies a broad class of deep learning architectures that operate on raw point clouds. 
It uses a shared multi-layer perceptron (MLP) to lift points to high-demensional space and then aggregates features of points using symmetric set function. PointNet++~\cite{qi2017pointnetplusplus} exploits local context by building hierarchical abstraction of point clouds. DGCNN~\cite{dgcnn} uses graph neural networks (GCN) on the $k$-nearest neighbor graphs to learn geometric features. KPConv~\cite{thomas2019kpconv} defines a set of kernel points to perform deformable convolutions, providing more flexibility than fixed grid convolutions. PCNN~\cite{pcnn2018} defines extension and restriction operations, mapping point cloud functions to volumetric functions and vice versa. SPLATNet~\cite{Su2018SPLATNetSL} renders point clouds to lattice grid and perform lattice convolutions.

FlowNet3D~\cite{flownet3d} and MeteorNet~\cite{Liu_2019_ICCV} adopt these methods and learn point-wise flows on dynamical point clouds. In addition to high-level point cloud recognition, recent works~\cite{Wang2019DCP,Wang2019PRNetSL,Goforth2019,Sarode2019OneFT} tackle low-level registration problems using point cloud networks and show significant improvements over traditional optimization-based methods. These point-based approaches, however, are constrained by the number of points in the point clouds and cannot scale to large-scale settings such as autonomous driving. To that end, sparse 3D convolutions \cite{sparseconv} have been proposed to apply 3D convolutions sparsely only on areas where points reside. Minkowski ConvNet~\cite{choy20194d} generalizes the definition of high-dimensional sparse convolution and improves 3D temporal perception. 

\textit{3D object detection.} The community has seen rising interest in 3D object detection thanks to the popularity of autonomous driving. VoxelNet~\cite{Zhou_2018_CVPR} proposes a generic one-stage model for 3D object detection. It voxelizes the whole point cloud and uses dense 3D convolutions to perform feature learning. To address the efficiency issue, PIXOR~\cite{Yang2018PIXORR3} and PointPillars~\cite{Lang_2019_CVPR} both organize in vertical columns (pillars); a PointNet is used to transform features from points to pillars. MVF~\cite{Zhou2019EndtoEndMF} learns to utilize the complementary information from both birds-eye view pillars and perspective view pillars. Complex-YOLO~\cite{Simon2018ComplexYOLOR3} extends YOLO to 3D scenarios and achieves real-time 3D perception; PointRCNN~\cite{shi2019pointrcnn}, on the other hand, adopts a RCNN-style detection pipeline. Rather than working in 3D, LaserNet~\cite{Meyer2019LaserNetAE} performs convolutions in raw range scans. 
Beyond point clouds only, recent works~\cite{ku2018joint,Chen2016Multiview3O,Xu2017PointFusionDS} combine point clouds with camera images to utilize additional information. Frustum-PointNet~\cite{qi2017frustum} leverages 2D object detectors to form a frustum crop of points and then uses a PointNet to aggregate features from points in frustum. \cite{Liang_2018_ECCV} designs an end-to-end learnable architecture that exploits continuous convolutions to have better fused feature maps in every level. In addition to visual inputs, \cite{yang18b} shows that High-Definition (HD) maps provide strong priors that can boost the performance of 3D object detectors. \cite{Liang_2019_CVPR} argues multi-tasking training can help the network to learn better representations than single-tasking. Beyond supervised learning, \cite{Wong2019IdentifyingUI} investigates how to learn a perception model for unknown classes. 
\section{Method}

In this section, we detail our approach to object pillar-based detection. 
We establish preliminaries about the pillar-point projection, PointPillars, and MVF in~\S\ref{sec:preliminaries} and summarize our model in~\S\ref{sec:archi}. Next, we discuss three critical new components of our model: cylindrical view projection (\S\ref{sec:cyv}), a pillar-based prediction paradigm (\S\ref{sec:pillar}), and a pillar-to-point projection module with bilinear interpolation (\S\ref{sec:method:bilinear}).  Finally, we introduce the loss function in~\S\ref{sec:loss}.  For ease of comparison to previous work, we use the same notation as MVF~\cite{Zhou2019EndtoEndMF}.

\subsection{Preliminaries}
\label{sec:preliminaries}
We consider a three-dimensional point cloud with $N$ points $P = \{p_i\}_{i=0}^{N-1} \subseteq \R^3$ with $K$-dimensional features $F = \{f_i\}_{i=0}^{N-1} \subseteq \R^K$. 
We define two functions $F_V(p_i)$ and $F_P(v_j)$,
where $F_V(p_i)$ returns the index $j$ of $p_i$'s corresponding pillar $v_j$ and $F_P(v_j)$ gives the set of  points in $v_j$. 
When projecting features from points to pillars, multiple points can potentially fall into the same pillar. To aggregate features from points in a pillar, a PointNet~\cite{qi2017pointnet} (denoted as $\mathrm{PN}$) 
is used to aggregate features from points to get pillar-wise features, where 
\begin{equation}\label{eq:pillarfeature}
     \begin{split}
        f^{\mathrm{pillar}}_j = \mathrm{PN}(\{f_i | \forall p_i \in F_P(v_j)\}).
     \end{split}
\end{equation}
Then, pillar-wise features are further transformed through an additional convolutional neural network (CNN), notated $\phi^{\mathrm{pillar}} = \Phi(f^{\mathrm{pillar}})$ where $\Phi$ denotes the CNN. To retrieve point-wise features from pillars, the pillar-to-point feature projection is given by 
\begin{equation}\label{eq:pillar2point}
     \begin{split}
        f^{\mathrm{point}}_i = f^{\mathrm{pillar}}_j \quad\mathrm{and}\quad \phi^{\mathrm{point}}_i = \phi^{\mathrm{pillar}}_j, \quad\mathrm{where}\quad j = F_V(p_i).
     \end{split}
\end{equation}
While PointPillars only considers birds-eye view pillars and makes predictions based on the birds-eye feature map, MVF also incorporates spherical pillars. Given a point $p_i = (x_i, y_i, z_i)$, its spherical coordinates $(\varphi_i, \theta_i, d_i)$ are defined via
\begin{equation}\label{eq:spv}
     \begin{split}
        \varphi_i = \arctan{\frac{y_i}{x_i}} \qquad \theta_i = \arccos{\frac{z_i}{d_i}} \qquad d_i = \sqrt{x^2_i + y^2_i + z^2_i}.
     \end{split}
\end{equation}
We can denote the established point-pillar transformations as $(F^{\mathrm{bev}}_V(p_i), F^{\mathrm{bev}}_P(v_j))$ and $(F^{\mathrm{spv}}_V(p_i), F^{\mathrm{spv}}_P(v_j))$ for the birds-eye view and the spherical view, respectively. In MVF, pillar-wise features are learned independently in two views; then the point-wise features are gathered from those views using Eq.~\ref{eq:pillar2point}. Next, the fused point-wise features are projected to birds-eye view again and embedded through a CNN as in PointPillars. 

The final detection head for both PointPillars and MVF is an anchor-based module. Anchors, parameterized by $(x^a, y^a, z^a, l^a, w^a, h^a, \theta^a)$, are densely placed in each cell of the final feature map. During pre-processing, an anchor is marked as ``positive'' if its intersection-over-union (IoU) with a ground-truth box is above a prescribed positive threshold, and ``negative'' if its IoU is below a negative threshold; otherwise, the anchor is excluded in the final loss computation.   

\subsection{Overall architecture}
\label{sec:archi}
\begin{figure*}[t]

\centering
\includegraphics[width=\linewidth]{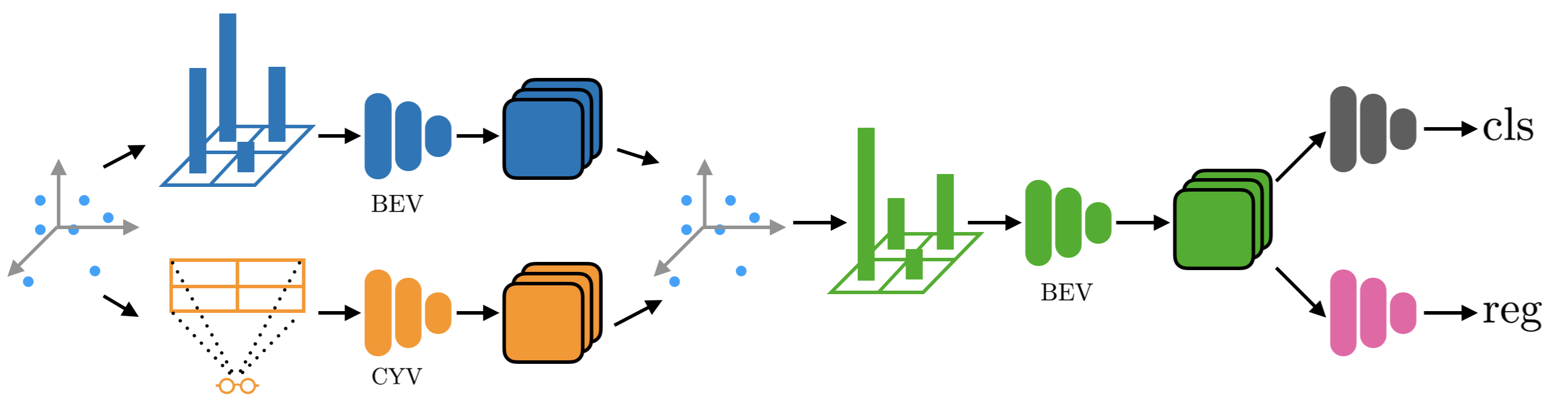}
\caption{Overall architecture of the proposed model: a point cloud is projected to BEV and CYV respectively; then, view-specific feature learning is done in each view; third, features from multiple views are aggregated; next, point-wise features are projected to BEV again for further embedding; finally, in BEV, a classification network and a regression network make predictions per pillar. BEV: birds-eye view; CYV: cylindrical view; cls: per pillar classification target; reg: per pillar regression target.}
\label{fig:archi}
\end{figure*}

An overview of our proposed model is shown in Figure~\ref{fig:archi}. The input point cloud is passed through the birds-eye view network and the cylindrical view network individually. Then, features from different views are aggregated in the same way with MVF. Next, features are projected back to birds-eye view and passed through additional convolutional layers. Finally, a classification network and a regression network make the final predictions per birds-eye view pillar. We \emph{do not} use anchors in any stage. We describe each module in detail below. 

\subsection{Cylindrical view}
\label{sec:cyv}

\begin{figure*}[t]

\centering
    \includegraphics[width=\linewidth]{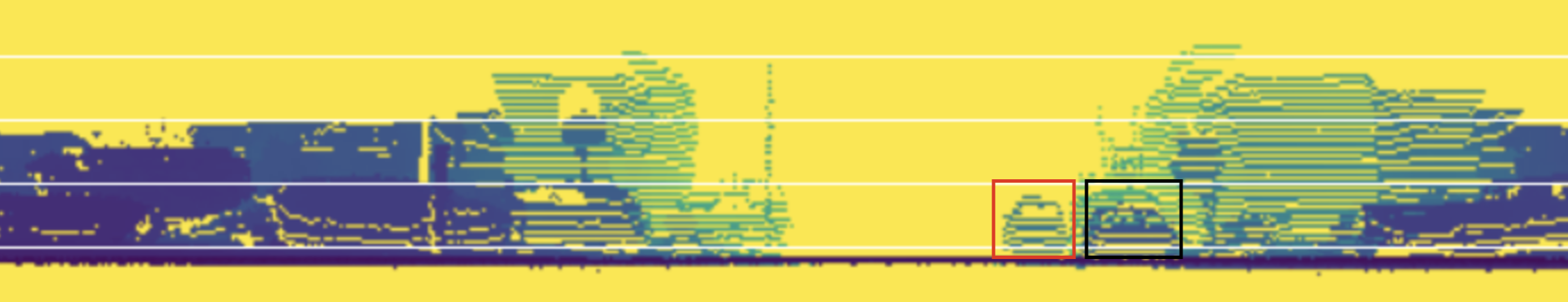}
    (a) Cylindrical View
    \includegraphics[width=\linewidth]{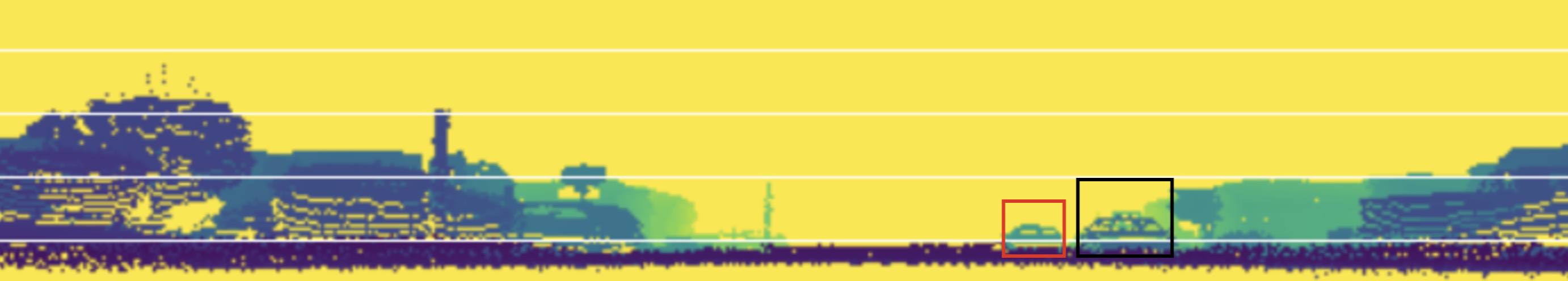}
    (b) Spherical View

\caption{Comparison of (a) cylindrical view projection and (b) spherical view projection. We label two example cars in these views. Objects in spherical view are distorted (in Z-axis) and no longer in physical scale.}
\label{fig:cyv}
\end{figure*}

In this section, we formulate the cylindrical view projection. 
The cylindrical coordinates ($\rho_i$, $\varphi_i$, $z_i$) of a point $p_i$ is given by the following:
\begin{equation}\label{eq:cyv}
     \begin{split}
        \rho_i = \sqrt{x^2_i+y^2_i} \quad\quad \varphi_i = \arctan{\frac{y_i}{x_i}} \quad\quad z_i = z_i. 
     \end{split}
\end{equation}
Cylindrical pillars are generated by grouping points that have the same $\varphi$ and $z$ coordinates. Although it is closely related to the spherical view, the cylindrical view does not introduce distortion in the Z-axis. We show an example in Figure~\ref{fig:cyv}, where cars are clearly visible in the cylindrical view but not distinguishable in the spherical view. In addition, objects in spherical view are no longer in their physical scales---e.g., distant cars become small. 

\subsection{Pillar-based prediction}
\label{sec:pillar}

\begin{figure*}[t]

\centering
    \begin{tabular}{@{}cc@{}}
    \includegraphics[height=1.5in]{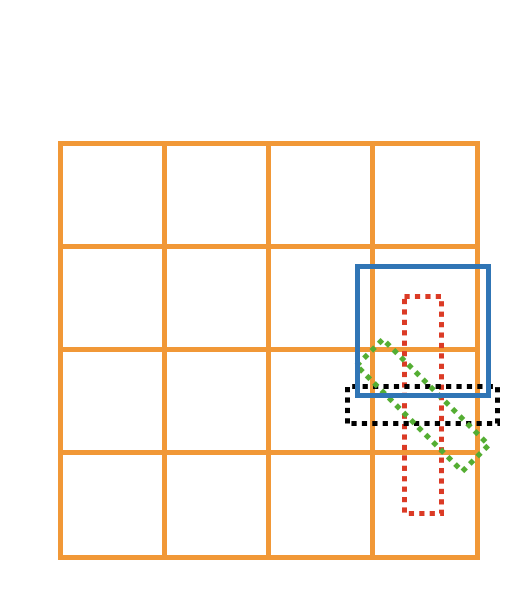}&
    \includegraphics[height=1.5in]{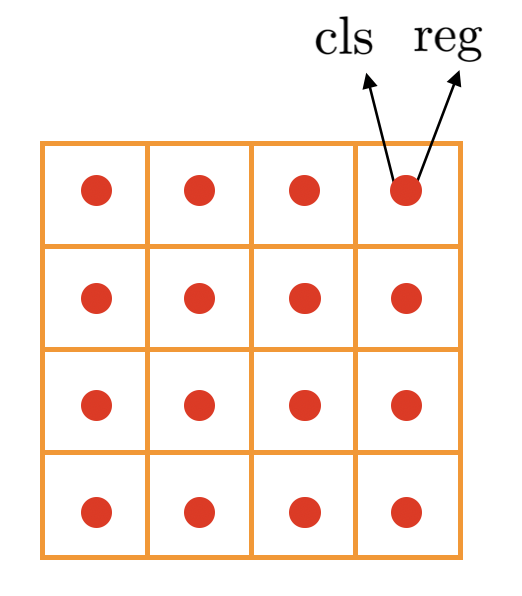} \\
  (a) Prediction per anchor & (b) Prediction per pillar
  \end{tabular}
  
\caption{Differences between prediction per anchor and prediction per pillar. (a) Multiple anchors with different sizes and rotations are densely placed in each cell. Anchor-based models predict parameters of bounding box for the positive anchor. For ease of visualization, we only show three anchors. Grid (in orange): birds-eye view pillar; dashed box (in red): a positive match; dashed box (in black): a negative match; dashed box (in green): invalid anchors because their IoUs are above negative threshold and below positive threshold. (b) For each pillar (center), we predict whether it is within a box and the box parameters. Dots (in red): pillar center.  
}
\label{fig:anchor:pillar}
\end{figure*}

The pillar-based prediction module consists of two networks: a classification network and a regression network. They both take the final pillar features $\phi^{\mathrm{pillar}}$ from birds-eye view. The prediction targets are given by
\begin{equation}\label{eq:pred}
     \begin{split}
        \mathrm{p} = f_{\mathrm{cls}}(\phi^{\mathrm{pillar}}) \quad\mathrm{and}\quad
        (\Delta_x, \Delta_y, \Delta_z, \Delta_l, \Delta_w, \Delta_h, \theta^p) = f_{\mathrm{reg}}(\phi^{\mathrm{pillar}}),
     \end{split}
\end{equation}
where p denotes the probability of whether a pillar is a positive match to a ground-truth box and ($\Delta_x, \Delta_y, \Delta_z, \Delta_l, \Delta_w, \Delta_h, \theta^p$) are the regression targets for position, size, and heading angle of the bounding box.  

The differences between anchor-based method and pillar-based method are explained in Figure~\ref{fig:anchor:pillar}. Rather than associating a pillar with an anchor and predicting the targets with reference to the anchor, the model (on the right) directly makes a prediction per pillar.

\subsection{Bilinear interpolation}
\label{sec:method:bilinear}

\begin{figure*}[t]

\centering
    \begin{tabular}{@{}cc@{}}
    \includegraphics[height=0.9in]{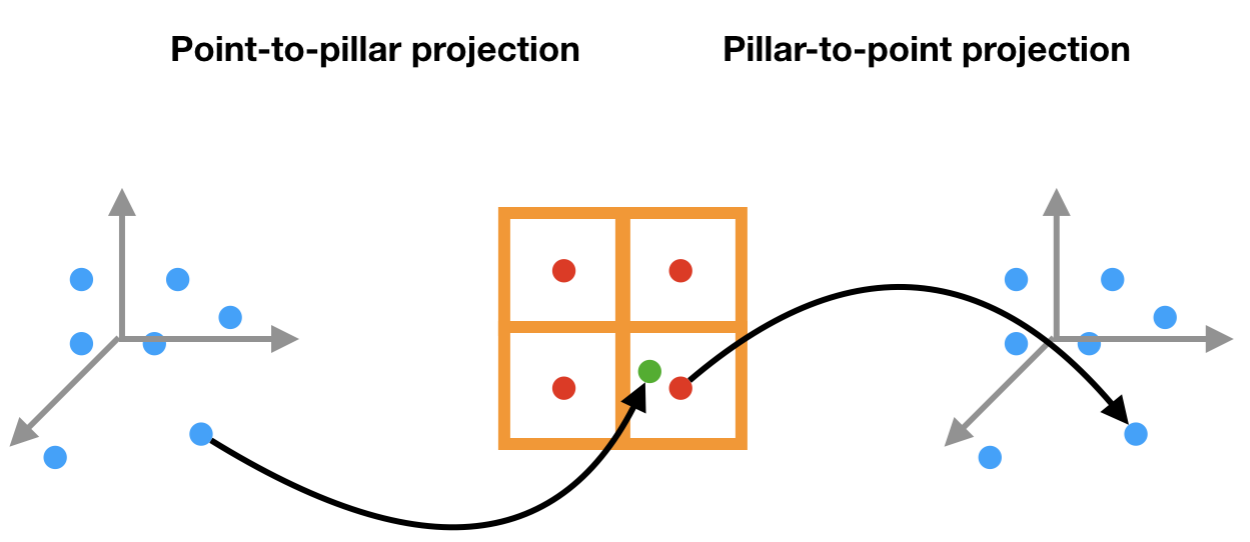}&
    \includegraphics[height=0.9in]{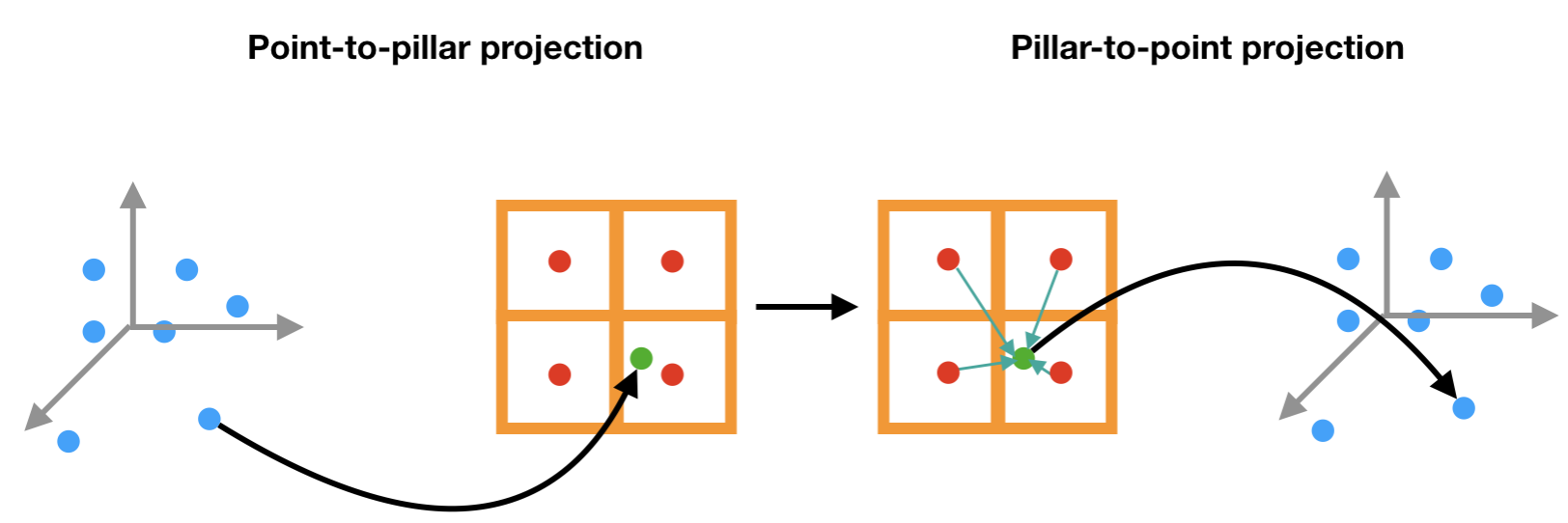} \\
  (a) Nearest neighbor interpolation & (b) Bilinear interpolation
  \end{tabular}
  
\caption{Comparison between nearest neighbor interpolation and bilinear interpolation in pillar-to-point projection. Rectangles (in orange): birds-eye view pillars; dots (in blue): points in 3D Cartesian coordinates; dots (in green): points projected to pillar frame; 
dots (in red): centers of pillars. }
\label{fig:bilinear}
\end{figure*}

The pillar-to-point projection used in PointPillars~\cite{Lang_2019_CVPR} and MVF~\cite{Zhou2019EndtoEndMF} can be thought of as a version of nearest neighbor interpolation, however, which often introduces quantization errors. Rather than performing nearest neighbor interpolation, we propose using bilinear interpolation to learn spatially-consistent features. We describe the formulation of nearest neighbor interpolation and bilinear interpolation in the context of pillar-to-point projection below. 


As shown in Figure~\ref{fig:bilinear} (a), we denote 
the center of a pillar $v_j$ as $c_j$ where $c_j$ is defined by its 2D or 3D coordinates. Then, the point-to-pillar mapping function is given by
\begin{equation}\label{eq:nearest}
     \begin{split}
        F_V(p_i) = j, \quad\textrm{where}\ \lVert p_i - c_j\rVert \leq \lVert p_i-c_k\rVert \quad \forall k
     \end{split}
\end{equation}
and $\lVert \cdot \rVert$ denotes the $\mathcal{L}_2$ norm. When querying the features for a point $p_i$ from a collection pillars, we determine the corresponding pillar $v_j$ by checking $F_V$ and copy the features of $v_j$ to $p_i$---that is $\phi^{\mathrm{point}}_i = \phi^{\mathrm{pillar}}_j$. 

This operation, though straightforward, leads to undesired spatial misalignment: if two points $p_i$ and $p_j$ with different spatial locations reside in the same pillar, their pillar-to-point features are the same. 
To address this issue, we propose using bilinear interpolation for the pillar-to-point projection. 
As shown in Figure~\ref{fig:bilinear} (b), the bilinear interpolation provides consistent spatial mapping between points and pillars.  


\subsection{Loss function}
\label{sec:loss}
We use the same loss function as in SECOND~\cite{Yan2018SECONDSE}, PointPillars~\cite{Lang_2019_CVPR}, and MVF~\cite{Zhou2019EndtoEndMF}. The loss function consists of two terms: a pillar classification loss and a pillar regression loss. 
The ground-truth bounding box is parametrized as $(x^g, y^g, z^g, \\ l^g, w^g, h^g, \theta^g)$; the center of pillar is $(x^p, y^p, z^p)$; and the prediction targets for the bounding box are ($\Delta_x, \Delta_y, \Delta_z, \Delta_l, \Delta_w, \Delta_h, \theta^p$) as in~\S\ref{sec:pillar}. Then, the regression loss is:
\begin{equation}\label{eq:loss}
     \begin{split}
        L_{\mathrm{reg}} & = \mathrm{SmoothL1}(\theta^p-\theta^g) + \sum_{r \in \{x, y, z\}} \mathrm{SmoothL1}(r^p-r^g-\Delta_r) \\ 
        & + \sum_{r \in \{l, w, h\}} \mathrm{SmoothL1}(\log(r^g)-\Delta_r)
     \end{split}
\end{equation}
where 
\begin{equation}\label{eq:smoothl1}
     \begin{split}
        \mathrm{SmoothL1}(d) = \left\{
                \begin{array}{ll}
                  0.5 \cdot d^2 \cdot \sigma^2, \quad\mathrm{if}\quad |d| < \frac{1}{\sigma^2} \\
                  |d| - \frac{1}{2\sigma^2}, \quad\mathrm{otherwise}.
                \end{array}
              \right.
     \end{split}
\end{equation}
We take $\sigma=3.0$. For pillar classification, we adopt the focal loss~\cite{lin2017focal}:
\begin{equation}\label{eq:focal}
     \begin{split}
        L_{\mathrm{cls}}  = -\alpha(1-\mathrm{p})^\gamma\log \mathrm{p}.
     \end{split}
\end{equation}
We use $\alpha=0.25$ and $\gamma=2$, as recommended by~\cite{lin2017focal}.

\section{Experiments}
\label{sec:exp}

Our experiments are divided into four parts. First, we demonstrate performance of our model for vehicle and pedestrian detection on the Waymo Open Dataset~\cite{Sun2019ScalabilityIP} in \S\ref{sec:waymo}. Then, we compare anchor-, point-, and pillar-based detection heads in \S\ref{sec:head}. We compare different combinations of views in \S\ref{sec:view}. Finally, we test the effects of bilinear interpolation in \S\ref{sec:bilinear}.  

\paragraph{Dataset.} The \textit{Waymo Open Dataset}~\cite{Sun2019ScalabilityIP} is the largest publicly-available 3D object detection dataset for autonomous driving. The dataset provides 1000 sequences total; each sequence contains roughly 200 frames. The training set consists of 798 sequences with 158,361 frames, containing 4.81M vehicle and 2.22M pedestrian boxes. The validation set consists of 202 sequences with 40,077 frames, containing 1.25M vehicle and 539K pedestrian boxes. The detection range is set to $[-75.2, 75.2]$ meters (m) 
horizontally and $[-3, 3]$ m vertically. 

\textit{Metrics.} For our experiments, we adopt the official evaluation protocols from the Waymo Open Dataset. In particular, we employ the 3D and BEV mean average precision (mAP) metrics. The orientation-aware IoU threshold is 0.7 for vehicles and 0.5 for pedestrians. We also break down the metrics according to the distances between the origin and ground-truth boxes: 0m--30m, 30m--50m, and 50m--infinity (Inf). The dataset is split based on the number of points in each box: LEVEL\_1 denotes boxes that have more than 5 points while LEVEL\_2 denotes boxes that have 1-5 points. Following StarNet~\cite{Ngiam2019StarNetTC}, MVF~\cite{Zhou2019EndtoEndMF}, and PointPillars~\cite{Lang_2019_CVPR} as reimplemented in~\cite{Sun2019ScalabilityIP}, we evaluate our models on LEVEL\_1 boxes.

\textit{Implementation details.} Our model consists of three parts: a multi-view feature learning network; a birds-eye view PointPillar~\cite{Lang_2019_CVPR} backbone; and a per-pillar prediction network. In the multi-view feature learning network, we project point features to both birds-eye view pillars and cylindrical pillars. For each view, we apply three ResNet~\cite{He2015} layers with strides $[1, 2, 2]$, which gradually 
downsamples the input feature to 1/1, 1/2, and 1/4 of the original feature map. Then, we project the pillar-wise features to points using bilinear interpolation and concatenate features from both views and from a parallel PointNet with one fully-connected layer. 
Then, we transform the per-point features to birds-eye pillars and use a PointPillars~\cite{Lang_2019_CVPR} backbone with three blocks to further improve the representations. The three blocks have $[4, 6, 6]$ convolutional layers, with dimensions $[128, 128, 256]$. Finally, for each pillar, the model predicts the categorical label using a classification head and 7 DoF parameters of its closest box using a regression head. The classification head and regression head both have four convolutional layers with 128 hidden dimensions. We use BatchNorm~\cite{Ioffe2015BatchNA} and ReLU~\cite{NairH10} after every convolutional layer. 

\textit{Training.} We use the Adam~\cite{Kingma2014AdamAM} optimizer to train the model. The learning rate is initially $3\times 10^{-4}$ and then linearly increased to $3\times10^{-3}$ in the first 5 epochs. Finally, the learning rate is decreased to $3\times10^{-6}$ using cosine scheduling~\cite{Loshchilov2017SGDRSG}. We train the model for 75 epochs in 64 TPU cores. 

\textit{Inference.} The input point clouds pass through the whole model once to get the initial predictions. Then, we use non-maximum suppression (NMS)~\cite{girshick15fastrcnn} to remove redundant bounding boxes. The oriented IoU threshold of NMS is 0.7 for vehicle and 0.2 for pedestrian. We keep the top 200 boxes for metric computation. The size of our model is on a par with MVF; the model runs at 15 frames per second (FPS) on a Tesla V100.

\subsection{Results compared to state-of-the-art}
\label{sec:waymo}

We compare the proposed method to top-performing methods on the Waymo Open Dataset. StarNet~\cite{Ngiam2019StarNetTC} is a purely point-based method with a small receptive field, which performs well for small objects such as pedestrians but poorly for large objects such as vehicles. LaserNet~\cite{Meyer2019LaserNetAE} operates on range images, which is similar to our cylindrical view feature learning. Although PointPillars~\cite{Lang_2019_CVPR} does not evaluate on this dataset, MVF~\cite{Zhou2019EndtoEndMF} and the Waymo Open Dataset~\cite{Sun2019ScalabilityIP} 
both re-implement the PointPillars. So we adopt the results from MVF~\cite{Zhou2019EndtoEndMF} and~\cite{Sun2019ScalabilityIP}. The re-implementation from~\cite{Sun2019ScalabilityIP} uses larger feature map resolution in the first PointPillars block; therefore, it outperforms the re-implementation from MVF~\cite{Zhou2019EndtoEndMF}. 

MVF~\cite{Zhou2019EndtoEndMF} extends PointPillars~\cite{Lang_2019_CVPR} with the same backbone networks and multi-view feature learning. We use the same backbone networks with PointPillars~\cite{Lang_2019_CVPR} and MVF~\cite{Zhou2019EndtoEndMF}. 

As shown in Table~\ref{table:car} and Table~\ref{table:ped}, we achieve significantly better results for both pedestrians and vehicles. Especially for distant vehicles (30m--Inf), the improvements are more significant.
This is inline with our hypothesis: in distant areas, anchors are less possible to  match to a ground-truth box; therefore, the imbalance problem is more serious. Also, to verify the improvements are \emph{not} due to differences in training protocol, we re-implement PointPillars; using our training protocol, it achieves 54.25 3D mAP and 70.59 2d mAP, which are worse than the re-implementations in~\cite{Zhou2019EndtoEndMF} and~\cite{Sun2019ScalabilityIP} . 
Therefore, we can conclude the improvements are due to the three new components added by our proposed model. 

\begin{table}[t] 
\begin{center}
\resizebox{\linewidth}{!}{
\begin{tabular}{c|c|c|c|c|c|c|c|c}
\toprule
\hline
\multirow{2}{*}{Method} & \multicolumn{4}{|c|}{BEV mAP (IoU=0.7)}  & \multicolumn{4}{|c}{3D mAP (IoU=0.7)} \\ 
\cline{2-9}
& Overall  &  0 - 30m  & 30 - 50m & 50m - Inf & Overall & 0 - 30m & 30 - 50m & 50m - Inf \\
\hline
StarNet~\cite{Ngiam2019StarNetTC} & - & - & - & - & 53.7 & - & - & - \\
\hline
LaserNet~\cite{Meyer2019LaserNetAE} & 71.57 & 92.94 & 74.92 & 48.87 & 55.1 & 84.9 & 53.11 & 23.92 \\
\hline
PointPillars$\P$~\cite{Lang_2019_CVPR} & 80.4 & 92.0 & 77.6 & 62.7 & 62.2 & 81.8 & 55.7 & 31.2 \\
\hline
PointPillars$\ddag$~\cite{Lang_2019_CVPR} & 70.59 & 86.63 & 69.34 & 49.3 & 54.25 & 76.31 & 48.08 & 24.21 \\
\hline
PointPillars$\dag$~\cite{Lang_2019_CVPR} & 75.57 & 92.1 & 74.06 & 55.47 & 56.62 & 81.01 & 51.75 & 27.94 \\
\hline
MVF~\cite{Zhou2019EndtoEndMF} & 80.4 & 93.59 & 79.21 & 63.09 & 62.93 & 86.3 & 60.2 & 36.02 \\
\hline
Ours  & \textbf{87.11} & \textbf{95.78} & \textbf{84.74} & \textbf{72.12} & \textbf{69.8} & \textbf{88.53} & \textbf{66.5} & \textbf{42.93} \\\hline
Improvements & \textbf{+6.71} & \textbf{+2.19} & \textbf{+5.53} & \textbf{+9.03} & \textbf{+6.87} & \textbf{+2.23} & \textbf{+6.3} & \textbf{+6.91} \\
\hline
\bottomrule
\end{tabular}
}
\end{center}
\vspace{-2pt}
\caption{Results on vehicle\label{table:car}. $\P$: re-implemented by~\cite{Sun2019ScalabilityIP}, the feature map in the first PointPillars block is two times as big as in others; $\ddag$: our re-implementation; $\dag$: re-implemented by~\cite{Zhou2019EndtoEndMF}.}
\end{table}

\begin{table}[t] 
\begin{center}
\resizebox{\linewidth}{!}{
\begin{tabular}{c|c|c|c|c|c|c|c|c}
\toprule
\hline
\multirow{2}{*}{Method} & \multicolumn{4}{|c|}{BEV mAP (IoU=0.5)}  & \multicolumn{4}{|c}{3D mAP (IoU=0.5)} \\ 
\cline{2-9}
& Overall  &  0 - 30m  & 30 - 50m & 50m - Inf & Overall & 0 - 30m & 30 - 50m & 50m - Inf \\
\hline
StarNet~\cite{Ngiam2019StarNetTC} & - & - & - & - & 66.8 & - & - & - \\
\hline
LaserNet~\cite{Meyer2019LaserNetAE} & 70.01 & 78.24 & 69.47 & 52.68 & 63.4 & 73.47 & 61.55 & 42.69\\
\hline
PointPillars$\P$~\cite{Lang_2019_CVPR} & 68.7 & 75.0 & 66.6 & 58.7 & 60.0 & 68.9 & 57.6 & 46.0 \\
\hline
PointPillars$\dag$~\cite{Lang_2019_CVPR} & 68.57 & 75.02 & 67.11 & 53.86 & 59.25 & 67.99 & 57.01 & 41.29 \\
\hline
MVF~\cite{Zhou2019EndtoEndMF} & 74.38 & 80.01 & 72.98 & 62.51 & 65.33 & 72.51 & 63.35 & 50.62 \\
\hline
Ours & \textbf{78.53} & \textbf{83.56} & \textbf{78.7} & \textbf{65.86} & \textbf{72.51} & \textbf{79.34} & \textbf{72.14} & \textbf{56.77} \\
\hline
Improvements & \textbf{+4.15} & \textbf{+3.55} & \textbf{+5.72} & \textbf{+3.35} & \textbf{+5.71} & \textbf{+6.83} & \textbf{+8.77} & \textbf{+6.15} \\
\hline
\bottomrule
\end{tabular}
}
\end{center}
\vspace{-2pt}
\caption{Results on pedestrian\label{table:ped}. $\P$: re-implemented by~\cite{Sun2019ScalabilityIP}. $\dag$: re-implemented by~\cite{Zhou2019EndtoEndMF}.}
\end{table}

\subsection{Comparing anchor-based, point-based, and pillar-based prediction}
\label{sec:head}

In this experiment, we compare to alternative means of making predictions: predicting box parameters per anchor or per point. For these three detection head choices, we use the same overall architecture with experiments in \S\ref{sec:waymo}. We conduct this ablation study on vehicle detection. 

\textit{Anchor-based model.} We use the parameters and matching strategy from PointPillars~\cite{Zhou2019EndtoEndMF} and MVF~\cite{Zhou2019EndtoEndMF}. Each class anchor is described by a width, length, height, and center position and is applied at two orientations: $0^\circ$ and $90^\circ$. Anchors are matched to ground-truth boxes using the 2D IoU with the following rules: a positive match is either the highest with a ground truth box, or above the positive match threshold (0.6); while a negative match is below the negative threshold (0.45). All other anchors are ignored in the box parameter prediction. The model is to predict whether a anchor is positive or negative, and width, length, height, heading angle, and center position of the bounding box. 

\textit{Point-based model.} The per-pillar features are projected to points using bilinear interpolation. Then, we assign each point to its surrounding box with the following rules: if a point is inside a bounding box, we assign it as a foreground point; otherwise it is a background point. The model is asked to predict the binary label whether a point is a foreground point or a background point. For positive points, the model also predicts the width, length, height, heading angle, and center offsets (with reference to point positions) of their associated bounding boxes. Conceptually, this point-based model is an instantiation of VoteNet~\cite{qi2019deep} applied to this autonomous driving scenario. The key difference is: the VoteNet~\cite{qi2019deep} uses a PointNet++~\cite{qi2017pointnetplusplus} backbone while we use a PointPillars~\cite{Zhou2019EndtoEndMF} backbone.

\textit{Pillar-based model.} Since we use the same architecture, we take the results from \S\ref{sec:waymo}. 
As Table~\ref{table:head} shows, anchor-based prediction performs the worst while point-based prediction is slightly better. Our pillar-based prediction is top performing among these three choices. The pillar-based prediction model achieves the best balance between coarse prediction (per anchor) and fine-grained prediction (per point). 

\begin{table}[t] 
\begin{center}
\resizebox{\linewidth}{!}{
\begin{tabular}{c|c|c|c|c|c|c|c|c}
\toprule
\hline
\multirow{2}{*}{Method} & \multicolumn{4}{|c|}{BEV mAP (IoU=0.7)}  & \multicolumn{4}{|c}{3D mAP (IoU=0.7)} \\ 
\cline{2-9}
& Overall  &  0 - 30m  & 30 - 50m & 50m - Inf & Overall & 0 - 30m & 30 - 50m & 50m - Inf \\
\hline
Anchor-based & 78.84 & 91.91 & 74.99 & 59.59 & 59.78 & 82.69 & 53.38 & 31.02 \\
\hline
Point-based & 79.77 & 92.35 & 76.58 & 60.00 & 60.6 & 83.66 & 55.48 & 30.95 \\
\hline
Pillar-based  & \textbf{87.11} & \textbf{95.78} & \textbf{84.74} & \textbf{72.12} & \textbf{69.8} & \textbf{88.53} & \textbf{66.5} & \textbf{42.93} \\
\hline
\bottomrule
\end{tabular}
}
\end{center}
\caption{Comparison of making prediction per anchor, per point, or per pillar.\label{table:head}}
\end{table}

\subsection{View combinations}
\label{sec:view}

In this section, we test different view projections in multi-view feature learning: birds-eye view (BEV), spherical view (SPV), XZ view, cylindrical view (CYV), and their combinations. 
First, we define the vehicle frame: the X-axis is positive forwards, the Y-axis is positive to the left, and the Z-axis is positive upwards. Then, we can write the coordinates of a point $p=(x, y, z)$ in different views; the range of each view is given in Table~\ref{table:project}. 
The pillars in the corresponding view are generated by projecting points from 3D to 2D using the coordinate transformation. One exception is in XZ view, in which we use separate pillars for positive part and negative part for Y-axis to avoid undesired occlusions. 

We show results of different view projections and their combinations in Table~\ref{table:view} for vehicle detection. When using a single view, the cylindrical view achieves significantly better results than the alternatives, especially in the long-range detection case (50m--Inf). When combining with the birds-eye view, the cylindrical view still outperforms others in all metrics. The spherical view, albeit similar to cylindrical view, introduces distortion in Z-axis, degrading performance relative to the cylindrical view. On the other hand, the XZ view does not distort the Z-axis, but occlusions in Y-axis prevent it from achieving as strong results as the cylindrical view. We also test with additional view combinations (such as using birds-eye view, spherical view, and cylindrical view) and do not observe any improvements over combining just the birds-eye view and the cylindrical view.

\begin{table}[t] 
\begin{center}
\resizebox{\linewidth}{!}{
\begin{tabular}{c|c|c}
\toprule
\hline
View & Coordinates & Range \\
\hline
3D Cartesian & (x, y, z) & (-75.2, 75.2)m, (-75.2, 75.2)m, (-3, 3)m \\
\hline
BEV & (x, y, z) & (-75.2, 75.2)m, (-75.2, 75.2)m, (-3, 3)m \\
\hline
SPV & ($\arctan(\frac{y}{x})$, $\arccos(\frac{z}{\sqrt{x^2+y^2+z^2}})$, $\sqrt{x^2+y^2+z^2}$) & (0, $2\pi$), (0.485$\pi$, 0.55$\pi$), (0, 107)m, \\ 
\hline
XZ view & (x, y, z) & (-75.2, 75.2)m , (-75.2, 75.2)m, (-3, 3)m\\ 
\hline
CYV & ($\sqrt{x^2+y^2}$, $\arctan(\frac{y}{x})$, z) & (0, 107)m, (0, 2$\pi$), (-3, 3)m  \\ 
\hline
\bottomrule
\end{tabular}
}
\end{center}
\caption{View projection~\label{table:project}}
\end{table}

\begin{table}[t] 
\begin{center}
\resizebox{\linewidth}{!}{
\begin{tabular}{c|c|c|c|c|c|c|c|c}
\toprule
\hline
\multirow{2}{*}{Method} & \multicolumn{4}{|c|}{BEV mAP (IoU=0.7)}  & \multicolumn{4}{|c}{3D mAP (IoU=0.7)} \\ 
\cline{2-9}
& Overall  &  0 - 30m  & 30 - 50m & 50m - Inf & Overall & 0 - 30m & 30 - 50m & 50m - Inf \\
\hline
BEV & 81.58 & 92.69 & 78.64 & 63.52 & 61.86 & 83.61 & 56.91 & 33.53 \\
\hline
SPV & 81.58 & 93.7 & 78.43 & 63.2 & 62.08 & 83.31 & 56.59 & 34.05 \\
\hline
XZ & 81.49 & 94.03 & 78.04 & 62.32 & 61.67 & 84.64 & 55.01 & 32.06 \\
\hline
CYV & 83.43 & 95.21 & 81.49 & 66.77 & 64.77 & 87.09 & 60.91 & 37.99 \\
\hline
BEV + SPV & 85.09 & 95.19 & 82.01 & 69.13 & 66.31 & 86.56 & 61.15 & 39.36 \\
\hline
BEV + XZ & 82.45 & 94.1 & 79.19 & 63.91 & 62.76 & 85.08 & 56.8 & 33.36 \\
\hline
BEV + CYV & \textbf{87.11} & \textbf{95.78} & \textbf{84.74} & \textbf{72.12} & \textbf{69.8} & \textbf{88.53} & \textbf{66.5} & \textbf{42.93} \\
\hline
\bottomrule
\end{tabular}
 }
\end{center}
\caption{Ablation on view combinations. \label{table:view}}
\end{table}

\subsection{Bilinear interpolation or nearest neighbor interpolation?}
\label{sec:bilinear}

In this section, we compare bilinear interpolation to nearest neighbor interpolation in pillar-to-point projection (for vechile detection). The architectures remain the same for both alternatives except the way we project multi-view features from pillars to points: In nearest neighbor interpolation, for each query point, we sample its closest pillar center and copy the pillar features to it, while in bilinear interpolation, we sample its four pillar neighbors and take a weighted average of the corresponding pillar features. Table~\ref{table:bilinear} shows bilinear interpolation systematically outperforms its counterpart in all metrics. This observation is consistent with the comparison of ROIAlign~\cite{he2017maskrcnn} and ROIPool~\cite{ren2015faster} in 2D. 

\begin{table}[t] 
\begin{center}
\resizebox{\linewidth}{!}{
\begin{tabular}{c|c|c|c|c|c|c|c|c}
\toprule
\hline
\multirow{2}{*}{Method} & \multicolumn{4}{|c|}{BEV mAP (IoU=0.7)}  & \multicolumn{4}{|c}{3D mAP (IoU=0.7)} \\ 
\cline{2-9}
& Overall  &  0 - 30m  & 30 - 50m & 50m - Inf & Overall & 0 - 30m & 30 - 50m & 50m - Inf \\
\hline
Nearest neighbor & 84.67 & 94.42 & 79.2 & 65.77 & 64.76 & 85.55 & 59.21 & 35.63 \\
\hline
Bilinear & \textbf{87.11} & \textbf{95.78} & \textbf{84.74} & \textbf{72.12} & \textbf{69.8} & \textbf{88.53} & \textbf{66.5} & \textbf{42.93} \\
\hline
\bottomrule
\end{tabular}
 }
\end{center}
\caption{Comparing bilinear interpolation and nearest neighbor projection. \label{table:bilinear}}
\end{table}
\section{Discussion}

We present a pillar-based object detection pipeline for autonomous driving. Our model achieves state-of-the-art results on the largest publicly-available 3D object detection dataset. The success of our model suggests many designs from 2D object detection/visual recognition are \emph{not} directly applicable to 3D scenarios. In addition, we find that learning features in correct views is import to the performance of the model.

Our experiments also suggest several avenues for future work. For example, rather than hand-designing a view projection as we do in~\S\ref{sec:cyv}, learning an optimal view transformation from data may provide further performance improvements. Learning features using 3D sparse convolutions rather than 2D convolutions could improve performance as well. Also, following two-stage object detection models designed for images, adding a refinement step might increase the performance for small objects.

Finally, we hope to find more applications of the proposed model beyond object detection. For example, we could incorporate instance segmentation, which may help with fine-grained 3D recognition and robotic manipulation.

\section{Acknowledgements}

Yue Wag, Justin Solomon, and the MIT Geometric Data Processing group acknowledge the generous support of Army Research Office grants W911NF1710068 and W911NF2010168, of Air Force Office of Scientific Research award FA9550-19-1-031, of National Science Foundation grant IIS-1838071, from the MIT--IBM Watson AI Laboratory, from the Toyota--CSAIL Joint Research Center, from gifts from Google and Adobe Systems, and from the Skoltech--MIT Next Generation Program. Any opinions, findings, and conclusions or recommendations expressed in this material are those of the authors and do not necessarily reflect the views of these organizations. 

\clearpage
%
%
\bibliographystyle{splncs04}
\bibliography{eccv2020}

\clearpage

\appendix
\section{Supplementary Material}
\label{sec:supplement}

\begin{figure*}[h!]
\centering
    \includegraphics[height=1.35in]{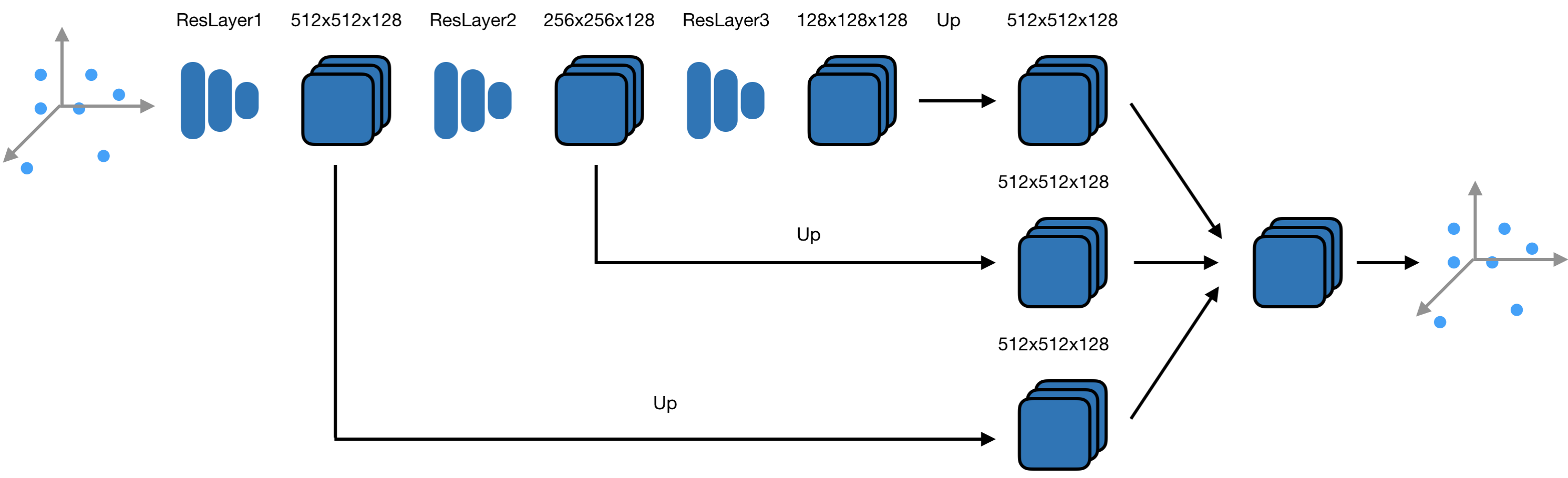} \\
    (a) Multi-view Feature Learning
    \begin{tabular}{@{}cc@{}}
    \includegraphics[height=1.1in]{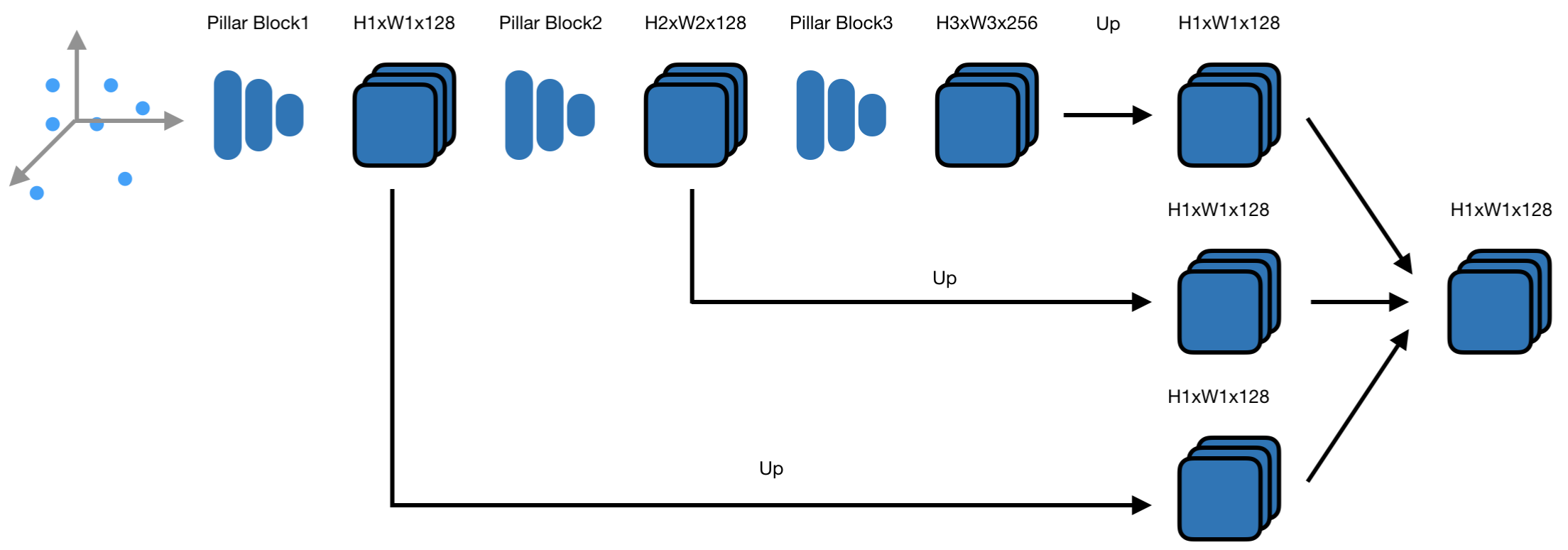}&
    \includegraphics[height=1.1in]{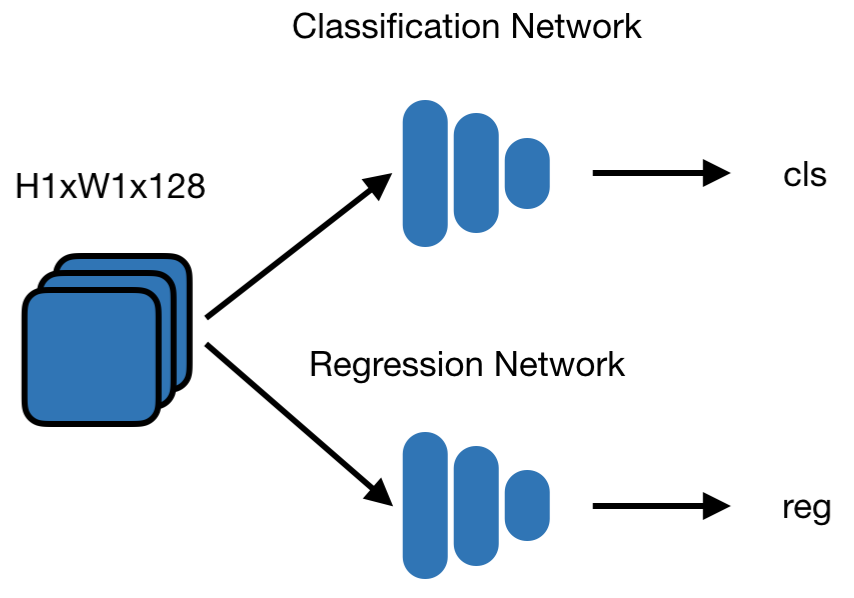} \\
  (b) Pillar Backbone & (c) Detection Head
  \end{tabular}
\caption{Details of the proposed model: (a) the multi-view feature learning module, we show the network for one view; (b) Pillar backbone network; (c) the detection head, we show both the classification network and the regression network. For details on the parameters and the feature map sizes, refer to Table~\ref{table:parameter}.}
\label{fig:network:detail}
\end{figure*}

In this section, we provide details on the parameters of the model. The model consists of  three parts: a multi-view feature learning network; a birds-eye view pillar backbone network; and a detection head. We show the pipeline in Figure~\ref{fig:network:detail} and the additional parameter specification in Table~\ref{table:parameter}.

\begin{table}
\begin{center}
\resizebox{\linewidth}{!}{
\begin{tabular}{c|c|c|c|c}
\toprule
\hline
\multirow{2}{*}{Stage} & \multicolumn{2}{|c|}{Vehicle Model}  & \multicolumn{2}{|c}{Pedestrian Model}  \\
 \cline{2-5}
 &  Kernel  & Output Size & Kernel & Output Size \\
\hline

\multirow{3}{*}{Multi-view Feature Learning}  &  3x3, 128, stride 1  & 512x512x128 & 3x3, 128, stride 1  & 512x512x128 \\
\cline{2-5}
 &  3x3, 128, stride 2  & 256x256x128 & 3x3, 128, stride 2  & 256x256x128 \\
\cline{2-5}
 &  3x3, 128, stride 2  & 128x128x128 &  3x3, 128, stride 2  & 128x128x128 \\
 
 \hline
 \multirow{2}{*}{Pillar Backbone Block1} &  3x3, 128, stride 2  & 256x256x128 & 3x3, 128, stride 1  & 512x512x128 \\
 \cline{2-5}
 &  \{3x3, 128, stride 1\}x3  & 256x256x128 & \{3x3, 128, stride 1\}x3  & 512x512x128 \\
 
 \hline
  \multirow{2}{*}{Pillar Backbone Block2} &  3x3, 128, stride 1  & 256x256x128 & 3x3, 128, stride 2  & 256x256x128 \\
 \cline{2-5}
 &  \{3x3, 128, stride 1\}x5  & 256x256x128 & \{3x3, 128, stride 1\}x5 & 256x256x128 \\
 
 \hline
   \multirow{2}{*}{Pillar Backbone Block3} &  3x3, 256, stride 2  & 128x128x256 & 3x3, 256, stride 2  & 128x128x256 \\
 \cline{2-5}
 &  \{3x3, 256, stride 1\}x5  & 128x128x256 & \{3x3, 256, stride 1\}x5  & 128x128x256 \\
 \hline
  Detection Head &  \{3x3, 256, stride 1\}x4  & 256x256x256 & \{3x3, 256, stride 1\}x4  & 512x512x256 \\
\hline
\bottomrule
\end{tabular}
}
\end{center}
\caption{Parameters of convolutional kernels and feature map sizes.}
\label{table:parameter}
\end{table}

\end{document}